\documentclass[conference]{IEEEtran}
\IEEEoverridecommandlockouts
\usepackage{cite}
\usepackage{amsmath,amssymb,amsfonts}
\usepackage{algorithmic}
\usepackage{graphicx}
\usepackage{textcomp}
\usepackage{xcolor}
\usepackage{hyperref}
\usepackage{booktabs}
\usepackage{multirow}
\usepackage{subcaption}
\usepackage{listings}
\usepackage{float}
\usepackage{array}
\usepackage{algorithm}
\usepackage{fancyvrb}
\usepackage{enumitem}
\usepackage{listings}
\usepackage{xcolor}
\usepackage{courier}
\usepackage{placeins}
\usepackage[most]{tcolorbox}

\definecolor{accent}{RGB}{0,90,150}      
\definecolor{accent2}{RGB}{0,128,96}     
\definecolor{codebg}{gray}{0.96}         
\definecolor{codeborder}{gray}{0.70}     

\lstdefinelanguage{json}{
  basicstyle=\ttfamily\small,
  showstringspaces=false,
  breaklines=true,
  morestring=[b]",
  stringstyle=\color{accent2},
  morecomment=[l]{//},
  morecomment=[s]{/*}{*/},
  commentstyle=\color{black!50},
}
\lstdefinestyle{jsonstyle}{
  language=json,
  backgroundcolor=\color{codebg},
  frame=single,
  framerule=0.5pt,
  rulecolor=\color{codeborder},
  numbers=none,
  xleftmargin=0pt, xrightmargin=0pt,
  columns=fullflexible
}

\newtcolorbox{promptcard}[1][]{
  breakable,
  colback=white,          
  colframe=codeborder,    
  coltext=black,          
  arc=2mm,
  boxsep=1.1ex,
  left=1.4ex,right=1.4ex,top=0pt,bottom=1ex,
  colbacktitle=accent,
  coltitle=white,
  fonttitle=\bfseries\sffamily\small,
  enhanced,
  #1
}

\newcommand{\promptsubtitle}[1]{%
  \par\vspace{1.5ex}
  {\sffamily\bfseries\color{accent}#1}\par
  \vspace{1.0ex}
}

\setlist[itemize]{topsep=3pt,itemsep=2pt,leftmargin=1.2em}

\lstdefinelanguage{RouterPrompt}{
  morekeywords={Question:,Tools:,Reasoning:,Answer:,QUERY:,TOOLS:},
  alsoletter={:},
  sensitive=false
}

\lstdefinestyle{prompt}{
  language=RouterPrompt,
  basicstyle=\ttfamily\footnotesize,
  frame=single,
  rulecolor=\color{black!20},
  backgroundcolor=\color{gray!5},
  breaklines=true,
  showstringspaces=false,
  tabsize=2,
  captionpos=b,
  columns=fullflexible,
  numbers=left,
  numberstyle=\tiny\color{gray!70},
  numbersep=8pt,
  keywordstyle=\bfseries\color{blue!60!black},
  literate={YES}{{\textcolor{green!50!black}{YES}}}3
           {NO}{{\textcolor{red!70!black}{NO}}}2
           {\#\#\#\#\#\#}{{{\bfseries\color{blue!60!black}\#\#\#\#\#\#}}}6
}

\def\BibTeX{{\rm B\kern-.05em{\sc i\kern-.025em b}\kern-.08em
    T\kern-.1667em\lower.7ex\hbox{E}\kern-.125emX}}

\begin{document}

\title{Adaptive Data Flywheel: Applying MAPE Control Loops to AI Agent Improvement\\
{\footnotesize \textsuperscript{}}
}

\author{
\IEEEauthorblockN{Aaditya Shukla, Sidney Knowles, Meenakshi Madugula, Dave Farris, Ryan Angilly, Santiago Pombo,\\
Anbang Xu, Lu An, Abhinav Balasubramanian, Tan Yu, Jiaxiang Ren, Rama Akkiraju}
\IEEEauthorblockA{\textit{NVIDIA Corporation} \\
Santa Clara, CA, USA \\
 \{aadityaramsh, anbangx, rakkiraju\}@nvidia.com}
}

\maketitle

\begin{abstract}
Enterprise AI agents must continuously adapt to maintain accuracy, reduce latency, and remain aligned with user needs. We present a practical implementation of a data flywheel in NVInfo AI, NVIDIA’s Mixture-of-Experts (MoE) Knowledge Assistant serving over 30,000 employees. By operationalizing a MAPE-driven data flywheel, we built a closed-loop system that systematically addresses failures in retrieval-augmented generation (RAG) pipelines and enables continuous learning.
Over a 3-month post-deployment period, we monitored feedback and collected 495 negative samples. Analysis revealed two major failure modes: routing errors (5.25\%) and query rephrasal errors (3.2\%). Using NVIDIA NeMo microservices, we implemented targeted improvements through fine-tuning. For routing, we replaced a Llama 3.1 70B model with a fine-tuned 8B variant, achieving 96\% accuracy, a 10× reduction in model size, and 70\% latency improvement. For query rephrasal, fine-tuning yielded a 3.7\% gain in accuracy and a 40\% latency reduction.
Our approach demonstrates how human-in-the-loop (HITL) feedback, when structured within a data flywheel, transforms enterprise AI agents into self-improving systems. Key learnings include approaches to ensure agent robustness despite limited user feedback, navigating privacy constraints, and executing staged rollouts in production. This work offers a repeatable blueprint for building robust, adaptive enterprise AI agents capable of learning from real-world usage at scale.

\end{abstract}


\section{Introduction}

Enterprise adoption of generative AI (GenAI) agents has accelerated rapidly, with applications ranging from knowledge retrieval to workflow automation. However, the performance of these systems often deteriorates post-deployment due to evolving user intent, domain drift, and the absence of systematic feedback integration. A central challenge in operationalizing such agents lies in enabling them to continuously adapt based on real-world usage patterns and user feedback, without requiring full-scale retraining or infrastructure overhauls.

While retrieval-augmented generation (RAG) pipelines and Mixture-of-Experts (MoE) architectures have improved the relevance and efficiency of enterprise AI agents, most production deployments remain static and reactive. Feedback mechanisms, if present, are frequently decoupled from the model improvement process. This disconnect results in stagnant accuracy, increasing latency, and declining user trust. There is a pressing need for closed-loop systems that can monitor agent performance, analyze failure modes, and execute targeted optimizations in a cost-efficient and privacy-aware manner.

In this work, we introduce a MAPE-based data flywheel framework that enables continuous learning in enterprise AI agents through a modular, feedback-driven pipeline. Adapted from self-adaptive control loops, this framework supports the deployment of agents that evolve incrementally over time. We apply this approach to NVIDIA's deployment of NVInfo AI, an internal Knowledge Assistant Agent that serves over 30,000 employees across diverse domains including engineering, operations, HR, and sales. NVInfo AI integrates user feedback with performance telemetry to identify actionable failure signals and execute targeted updates using parameter-efficient fine-tuning (PEFT) and model specialization.

Over a three-month observation window, we collected and analyzed 495 negative feedback samples, revealing two dominant sources of failure: routing errors (5.25 \%) and query rephrasal errors (3.2 \%). Utilizing NVIDIA NeMo microservices, we applied lightweight, component-specific fine-tuning strategies to improve performance.

\begin{itemize}
    \item For routing, we reduced model size by a factor of ten (from 70 billion to 8 billion parameters) while maintaining 96\% accuracy and reducing latency by 70\%.

    \item For query rephrasal, we achieved a 3.7\% improvement in accuracy (measured on a synthetic dataset generated from manually analyzed incorrect queries, expanded to 5,000 examples and split 80/10/10), along with a 40\% reduction in response latency.

\end{itemize}

This work makes three key contributions:
\begin{itemize}
    \item We demonstrate \textbf{a novel application of the MAPE control loop to the domain of GenAI agent improvement}, bridging observability and action in a continuous feedback pipeline. 
    \item We present \textbf{an empirical analysis of post-deployment failure modes} in a production-grade enterprise AI agent, informed by real user feedback.
    \item We provide \textbf{a modular implementation blueprint using NVIDIA NeMo microservices}, offering a practical architecture for organizations seeking to build adaptive and self-correcting AI agents.
\end{itemize}

\section{Background and Related Work}

\subsection{From MAPE-K to Agentic AI: Foundations of Self-Adaptive Systems}
The MAPE-K (Monitor, Analyze, Plan, Execute – Knowledge) reference model, introduced by IBM \cite{ibm2006autonomic}, remains foundational for designing self-adaptive software systems by structuring behavior into a control loop that continuously responds to environmental changes, with its modular architecture enabling broad adoption across multiple domains \cite{iglesia2015mapek, arcaini2015modeling, rutten2017feedback, romero2022graph, andersson2023decentralized}. Central to its evolution is the Knowledge component, which supports long-term reasoning and intelligent adaptation, especially when integrated with machine learning to enable predictive and causal decision-making \cite{gheibi2021mlselfadaptive, abdennadher2022daacs, belhaj2018autonomic}. Within agentic AI frameworks, MAPE-K cycles are increasingly embedded in autonomous agents to drive real-time, decentralized adaptation that enables intelligent decision-making and responsive behavior in dynamic environments \cite{ patel2025agentic, hrabia2018adaptive}. As reinforcement learning and GenAI capabilities are incorporated into these loops, agents gain the ability to synthesize adaptive strategies and reason across modalities \cite{li2024generative}. These advancements illustrate MAPE-K’s synergy with the data flywheel paradigm: each monitoring cycle enriches the knowledge base, fueling increasingly effective planning and adaptation through a self-reinforcing loop \cite{sanwouo2025aware}.

\subsection{Modular Pipelines for Scalable RAG: Retrieval, Routing, and Rephrasal}
As enterprises adopt Retrieval-Augmented Generation (RAG) through staged pipelines that involve retrieval, grounding, reasoning, and feedback, they increasingly realize its value for scalability, compliance, and trustworthy AI, positioning RAG as a core enabler of intelligent adaptive systems \cite{akkiraju2024facts}. By grounding large language models in enterprise knowledge and facilitating continuous real-world feedback, RAG complements the MAPE-K trajectory and reinforces the data flywheel paradigm \cite{microsoft2024arena,nvidia2025nemo}. To support enterprise RAG deployments at scale, expert routing has emerged as a key architectural strategy for enabling adaptive, modular reasoning. Modular approaches such as Mixture of Experts (MoE) and multi-agent systems \cite{cai2025survey,zhou2022mixture} dynamically direct inputs to specialized components using techniques like embedding selectors, symbolic routing, and LLM-as-a-Router, enhancing efficiency and task-specific alignment \cite{chen2025symbolic,chen2024routerdc}. A complementary strand of research focuses on query understanding and rephrasal, which are critical for strengthening RAG pipelines. By mitigating ambiguity and poorly structured queries, these methods enhance retrieval accuracy and reduce hallucinations \cite{li2025unirag,dong2025leveraging}. Recent advances leverage LLM-based rephrasal, semantic parsing, and uncertainty-aware frameworks such as RaFe and Omni-RAG to clarify intent, improve retrievability, and boost response reliability in enterprise contexts \cite{mao2024rafe,yang2024rephrase,shrivastava2022retrieve}.

As these pipelines scale, the choice of model architecture becomes critical for balancing performance and cost. While 8B models offer lower latency and cost, they typically underperform compared to 70B models \cite{myscale2024llama,aws2024llama,bentoml2023benchmarking}. Parameter-efficient fine-tuning methods such as Low-Rank Adaptation (LoRA) and Quantized Low-Rank Adaptation (QLoRA) narrow this gap by adapting smaller models to specific tasks with minimal overhead, achieving near-parity in performance while reducing memory and compute demands by up to 100$\times$. This enables up to 60--80\% savings in GPU costs without compromising accuracy \cite{hu2022lora,dettmers2023qlora,coleman2025parameter,kim2025datacenter}.

\subsection{Feedback and Evaluation: Closing the Adaptation Loop}
To close the loop in these adaptive systems, human-in-the-loop (HITL) pipelines serve as a critical counterpart to automated feedback mechanisms. By embedding human expertise into monitoring, annotation, and evaluation stages, HITL workflows enhance the reliability and contextual accuracy of enterprise RAG deployments, particularly in high-stakes domains where model errors can lead to significant consequences \cite{vats2024survey,gama2014survey}. Modern approaches integrate subject matter experts, active learning, weak supervision, and toolkits such as Snorkel, Label Studio, and Prodigy to reduce annotation effort while enabling scalable and domain-aligned feedback cycles that continuously refine model behavior \cite{ratner2017snorkel,quotient2024smell,gong2024training}.

Robust evaluation plays a vital role in transforming HITL and system-generated feedback into actionable signals for model refinement. As a key driver of the data flywheel, it determines which behaviors to reinforce, retrain, or discard. Beyond traditional metrics like accuracy and latency, emerging methods such as LLM-as-a-Judge, reward modeling, and preference-based scoring more effectively capture alignment and robustness \cite{laskar2024survey,gao2025llm,tan2024judgebench,frick2024reward,zheng2023mtbench}.

\subsection{MAPE-K-Aligned Data Flywheel for Self-Improving Enterprise GenAI Systems}

Despite significant advances in retrieval, expert routing, rephrasal, fine-tuning, HITL feedback, and evaluation, enterprise GenAI systems often lack a cohesive architecture to support continuous adaptation. These components are typically implemented in isolation, limiting coordination between observability, retraining, and evaluation workflows. This fragmentation hampers responsiveness and diminishes reliability in production environments.

This paper presents \textbf{the first comprehensive application of MAPE-K principles to the improvement of AI agents in enterprise settings}. We introduce a MAPE-K-aligned data flywheel architecture that consolidates monitoring, analysis, planning, and execution into a modular, closed-loop pipeline. Leveraging NVIDIA’s NeMo Microservices \cite{nvidia2025nemobuild,nvidiadocs2025evaluate,nvidiadocs2025overview,constellation2025}, our framework integrates observability, feedback ingestion, fine-tuning, and evaluation, and supports secure, low-latency deployment across cloud, on-premises, and hybrid environments with built-in policy enforcement and real-time feedback handling \cite{unit82024,bitrock2024}.

By applying control-theoretic foundations to retrieval-augmented, multi-agent GenAI systems, our approach enables dynamic, self-improving behavior in production. The data flywheel continuously refines the knowledge base through each cycle of monitoring and evaluation, guiding targeted adaptations over time. This architecture provides a scalable and reliable foundation for building enterprise GenAI systems that evolve with real-world usage.

\section{System Architecture}

\subsection{NVInfo AI: Mixture of Experts Architecture}

Before describing the Adaptive Data Flywheel, we first present the underlying AI system it enhances. The NVInfo AI system operates as NVIDIA's internal enterprise chatbot which provides services to more than 30,000 staff members spread across different locations worldwide. The system operates with an advanced Mixture of Experts (MoE) framework which optimizes its performance when processing various enterprise information requests.

\subsubsection{Architecture Components}

The NVInfo AI system consists of multiple essential components (Figure \ref{fig:nvinfo_architecture}, Appendix \ref{sec:nvinfo_architecture}) which work together to generate precise answers that understand user context.

\begin{itemize}
    \item \textbf{User Interface}: The intranet portal functions as the main access point which allows staff members to ask questions and handles complex business information requirements across various domains. The system offers
        \begin{itemize}
            \item User questions through natural language while maintaining context understanding 
            \item Response Generation in table, lists and formatted data structure
            \item Source references which link directly to SharePoint documentation
            \item Follow-up question suggestions generated from conversational context
            \item Feedback system which uses thumbs up/down buttons to help agents improve their performance.
        \end{itemize}
    
    \item \textbf{Router Module}: The system uses Llama 3.1 70B as its initial large language model to classify user queries which then get sent to one of seven specialized experts.
        \begin{itemize}
            \item Financial Info Expert (earnings reports, transcripts)
            \item IT Help \& HR Benefits Expert (ServiceNow knowledge and catalog)
            \item SharePoint Expert (intranet content)
            \item Holidays Expert (region-specific holiday calendars)
            \item Cafe Menu Expert (cafeteria information)
            \item People Expert (organization charts, reporting chains)
            \item NVIDIA Policies Expert (corporate policies and procedures)
        \end{itemize}
    
    \item \textbf{Query Processing Pipeline}: The system processes queries through multiple stages after they pass through the router module.
        \begin{enumerate}
            \item Conversation Rephrasing: Incorporates prior turns for multi-turn dialogue.
            \item Query Variations: Generates multiple rephrasings to improve retrieval coverage.
            \item Retriever: Conducts semantic document searches across all available document collections.
            \item Re-ranking \& De-duplication: Ranks documents based on their relevance while removing duplicate results.
            \item Answer Generation: Creates a unified response by processing the retrieved information.
            \item Citation Generation: Produces trustworthy source links which enable users to verify information sources.
            \item Suggested Follow-ups: Generates additional questions which help users discover new content while enhancing their interaction with the system.
        \end{enumerate}
\end{itemize}

\subsubsection{NVInfo AI Conversation and Feedback Collection}
NVInfo AI system responds to user inquiries and simultaneously
records extensive conversational data and user feedback information.
The system records detailed response metrics and structured feedback metrics which get processed through a single data pipeline for system monitoring, evaluation and performance enhancement (Figure \ref{fig:nvinfo_data_capture}, Appendix \ref{sec:nvinfo_data_capture}).
\begin{itemize}
    \item \textbf{User Interaction}: The NVInfo User Interface enables Users to initiate queries. The NVInfo Agent receives input from the interface to execute query interpretation, information retrieval and response generation.
    \item \textbf{Response Metrics Collection}: The system generates NVInfo Response Metrics for every response it produces to enable observability and future analysis. The system tracks the following information points
        \begin{itemize}
            \item Query – the original user input
            \item Response – the agent’s output
            \item Category – the knowledge source from which information was retrieved
            \item Expert Selected – subject-matter expert or expert route chosen
            \item Time Taken – latency observed across different components in the agentic AI workflow
            \item Agent Thought – reasoning trace behind the response
            \item Rephrased Query – any reformulation of the user’s input
            \item IR Results – intermediate retrieval results
            \item Prompts – the prompt(s) used in response generation
            \item Guardrail Metrics – policy or safety checks applied to the response
        \end{itemize}
    These metrics are stored in DynamoDB, allowing fast, scalable storage of large volumes of response data.
    \item \textbf{Feedback Metrics Collection}: Users can give direct feedback through the thumbs up / thumbs down system after reviewing their responses. Users can access a feedback modal through these icons which allows them to add more details about their feedback. The system records the following NVInfo Feedback Metrics:
        \begin{itemize}
            \item Positive or negative signal (thumbs up/down)
            \item Contextual reasons for feedback, such as:
                 \begin{itemize}
                    \item Usefulness of cited sources
                    \item Relevance of the generated response
                    \item Clarity and completeness of the output
                    \item Suggestions for improvement
                \end{itemize}   
        \end{itemize}
    These structured feedback metrics are stored in a SQL Database, making them easy to query for analytics and fine-grained error analysis.
    \item \textbf{Unified Data Pipeline}: The data ingestion pipeline receives both response metrics from DynamoDB and feedback metrics from the SQL Database. The data transformation process within this pipeline creates a uniform schema structure while adding supplementary information and establishing signal type connections to achieve system interaction understanding. The processed data moves to a central Data Lake where it becomes available for comprehensive analysis, continuous monitoring, and iterative improvements to the NVInfo AI system.
\end{itemize}

\subsubsection{Performance Characteristics}

The baseline NVInfo AI operated with the following system metrics before Data Flywheel implementation:
\begin{itemize}
    \item Average response time: $\,\sim\,$12 seconds per query
    \item LLM as judge ratings: 4.2 correctness score out of 5 measured on our regression dataset (see Appendix \ref{sec:regression_data})
    \item Weekly query volume: $\,\sim\,$2000 unique queries across 800 unique users
\end{itemize}

This Mixture of Experts framework serves as the base structure which our Adaptive Data Flywheel system uses to enhance particular experts through user feedback analysis.

\begin{figure*}[!t]
\centering
\includegraphics[width=0.75\textwidth]{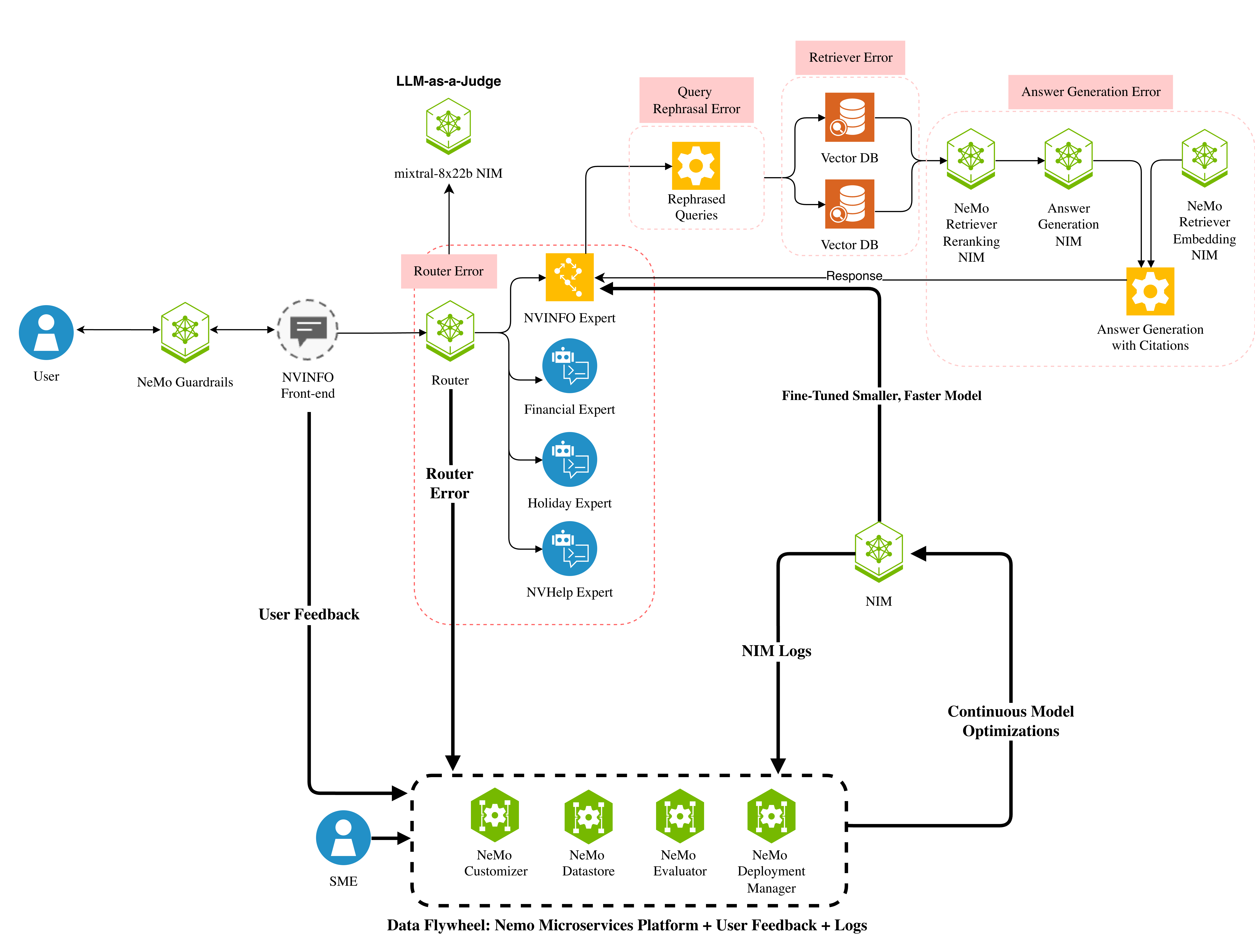}
\caption{Adaptive Data Flywheel Architecture showing the MAPE control loop implementation for AI agent improvement}
\label{fig:mape_architecture}
\end{figure*}

\subsubsection{RAG System Challenges}

Before introducing our solution, it is important to understand the failure points inherent in RAG-based systems. Figure \ref{fig:rag_failures} illustrates the failure modes observed in production, ordered by their occurrence in the processing pipeline (see Appendix \ref{sec:rag_failures}):

The RAG pipeline encounters multiple processing challenges throughout its entire operation:
\begin{enumerate}
    \item \textbf{Router - Query Understanding}: Misclassification of user intent leading to wrong expert selection. Example: "vacation days" queries routed to Holiday Expert instead of Policies Expert (5.25\% of our failures).
    
    \item \textbf{Query Rephrasing Error}: Incorrect expansion or interpretation of queries for the selected agent. Example: "RESS planning team" incorrectly rephrased as "Resource Planning team" instead of "Real Estate \& Site Services" (3.2\% of failures).
    
    \item \textbf{Retriever Error}: Failure to find relevant documents which exist in the knowledge base because of semantic search limitations or embedding mismatches.
    
    \item \textbf{Reranking Error}: Retrieved documents incorrectly prioritized which results in important information being hidden beyond the context window threshold.
    
    \item \textbf{LLM Hallucination}: The model produces believable yet false information when it lacks sufficient context which leads to confident but incorrect responses.
    
    \item \textbf{Citation Generation Error}: Incorrect or missing source references which decreases answer reliability and blocks users from verifying the information.
    
    \item \textbf{Answer Generation Error}: A poor final response by combining retrieved context which results in incomplete or unclear answers even though it has access to correct information.
\end{enumerate}

The MAPE control loop of our Adaptive Data Flywheel system solves these problems by implementing automatic detection and correction of these system failures.

\subsection{Adaptive Data Flywheel Overview}

Building upon the NVInfo AI architecture and addressing RAG system challenges, Figure \ref{fig:mape_architecture} illustrates how our Adaptive Data Flywheel wraps around the core system to enable continuous improvement. The flywheel contains the four MAPE phases with dedicated components for AI agent management which operate through a unified knowledge base.

\subsubsection{Monitor Component (\textbf{M} in MAPE)}
\begin{enumerate}[label=\textit{\Roman*.}, leftmargin=3em]
    \item \textit{Problem}: The numerous failure points in RAG pipelines make it challenging to identify between situational and systemic problems. Systemic problems remain undetected until users file complaints which results in negative user experiences and delayed solution implementation.
    \item \textit{Solution}: We implemented a comprehensive monitoring system that tracks both direct user feedback (thumbs up/down) and implicit signals (re-queries, session abandonment). Table \ref{tab:monitor_errors} shows two examples out of 495 queries that users gave a "thumbs-down" in the first three months after release. These queries demonstrate how two specific data points could be generalized into larger patterns for developers to address.
    \item \textit{Challenges}: 
    The collection of user feedback in AI systems encounters multiple obstacles which reduce the quality of evaluation signals. The main problem stems from insufficient user engagement because users only provide feedback to a limited extent which results in unrepresentative data collection. Users tend to report negative feedback more frequently because they focus on sharing their dissatisfaction with unsatisfactory results instead of verifying positive outcomes. The process of data collection becomes harder because organizations need to remove all personally identifiable information (PII) from query–response pairs while following strict privacy and security regulations. The available feedback data remains incomplete because users mainly provide basic binary feedback such as thumbs up or down without explaining their reasons for dissatisfaction. Users sometimes provide incorrect feedback through comments that fail to pinpoint the actual cause of their negative ratings. The combination of these factors makes it difficult to obtain dependable feedback that can help systems improve their performance.

    \item \textit{Learnings}: 
        User feedback collection needs systems that combine user-friendly interfaces with privacy protection features to achieve better participation rates and useful data insights. The system should allow users to rate content directly and complete brief surveys and follow step-by-step prompts for feedback collection while maintaining full compliance with GDPR and CCPA regulations and enterprise data protection policies that ban personal information storage. The system should record both direct ratings and indirect feedback indicators which include user interaction statistics, search pattern changes and repeated query attempts. The system needs to request positive feedback through confirmation questions (e.g. “Was this answer helpful?”) together with negative feedback collection methods. Real-time feedback processing systems need to be established for immediate input evaluation and classification to shorten model improvement cycles.
\end{enumerate}

\begin{table*}[!t]
\centering
\caption{Representative Error Examples Captured by Monitor Component During 3-Month Deployment}
\label{tab:monitor_errors}
\begin{tabular}{p{4cm}p{6cm}p{3cm}p{3cm}}
\toprule
\textbf{User Query} & \textbf{System Response/Issue} & \textbf{Error Type} & \textbf{Impact} \\
\midrule
"What is the role of the RESS planning team at NVIDIA?" & Unable to find answer - RESS incorrectly expanded to "Resource Planning team" instead of "Real Estate \& Site Services" & Query Rephrasing & Failed to retrieve correct department information \\
\midrule
"How many vacation days does NVIDIA Canada have?" & "I don't have enough information to answer this question" & Router Error & Sent to Holiday Expert instead of Policies Expert \\
\bottomrule
\end{tabular}
\end{table*}

\subsubsection{Analyze Component (\textbf{A} in MAPE)}
\begin{enumerate}[label=\textit{\Roman*.}, leftmargin=3em]
    \item \textit{Problem}: Raw feedback data tends to lack actionable insights. The RAG pipeline contains multiple failure points (see Figure \ref{fig:rag_failures}) which makes it difficult to identify original causes and determine which components caused the errors. Without accurate error attribution, developers may introduce fixes that fail to significantly improve answer quality.
    \item \textit{Solution}: We developed systematic error attribution techniques combining manual analysis with automated classification. From 495 thumbs-down samples:
        \begin{itemize}
            \item \textbf{Routing Errors}: 26/495 (5.25\%) - Queries sent to wrong expert
            \item \textbf{Rephrasal Errors}: $\sim$3.2\% (extrapolated from analyzing 250/495 samples)
        \end{itemize}
    Although the NVInfo expert routing classifier demonstrated high overall accuracy, our analysis revealed that certain low-frequency query classes exhibited poor data representation. This distributional imbalance led to occasional misclassifications within those specific subsets. Recognizing this gap, we designed targeted experiments to enrich the data and improve performance in those underrepresented domains.
    Specific examples identified:
        \begin{itemize}
            \item \textbf{Routing Error}: "How many vacation days does NVIDIA Canada have?" was sent to NVINFO's Holiday Expert instead of the Policies Expert
            \item \textbf{Rephrasal Error}: "RESS planning team" incorrectly rephrased as "NVIDIA Resource Planning team" instead of "Real Estate \& Site Services"
        \end{itemize}
    \item \textit{Challenges}: 
       The RAG pipeline contains multiple failure points throughout its different stages as shown in Section III. The propagation of initial routing mistakes through subsequent components leads to cascading errors which grow more severe with each stage. The process of manual analysis creates a bottleneck because expert review is needed to perform accurate attribution. The identification of root causes becomes difficult when issues present as ambiguous failures because multiple dependent factors create the overall error.
    \item \textit{Learnings}: 
        The RAG pipeline needs tracing functionality to track queries, retrieval operations and model choices because this will help developers debug the system efficiently and identify where failures occur. The attribution models which use heuristics or machine learning classifiers help identify which stages of the pipeline produce errors. The system needs to distinguish between model-related breakdowns and non-model problems because this separation enables developers to identify LLM-related errors from retrieval and ranking system errors. The evaluation of different system configurations (chunking methods and embedding models) through A/B testing will show their individual performance effects. The process of error classification and root-cause identification becomes faster through automated issue labeling which uses weak supervision or heuristic tagging methods.     
\end{enumerate}

\subsubsection{Plan Component (\textbf{P} in MAPE)}
\begin{enumerate}[label=\textit{\Roman*.}, leftmargin=3em]
    \item \textit{Problem}: The developers need to make extensive modifications across multiple system components to fix the fundamental problems they have discovered. The combination of restricted labeled data, privacy restrictions and specialized domain requirements makes standard model retraining methods ineffective.
    \item \textit{Solution}: We developed targeted data curation and fine-tuning strategies for different problems leveraging NVIDIA NeMo microservices: \\
    \textbf{Routing Error Remediation:}
        \begin{itemize}
            \item Collected user feedback + SME corrected completions
            \item Used LLM-as-a-Judge: 140 incorrect routing identified, 32 truly incorrect
            \item Created ground truth: 761 data points (729 original + 32 corrections)
            \item Final dataset: 685 samples after removing duplicates (60/40 train/test split)
        \end{itemize}
    \textbf{Rephrasal Error Remediation:}
        \begin{itemize}
            \item Collected user feedback and manually analyzed 250/495 thumbs-down samples
            \item 10 incorrect rephrasal identified, shortlisted 4
            \item Synthetic data generation: Generated 5,000 synthetic samples using 4 shortlisted examples as few-shot prompts to Llama 3.1 405B (see Appendix \ref{sec:synthetic_data_generation})
            \item Final dataset: 5,000 synthetic data samples (80/10/10 train/validation/test split)
        \end{itemize}
    \textbf{Implementation Tools:}
        \begin{itemize}
            \item \textbf{NeMo Curator}: Data processing and cleaning
            \item \textbf{NeMo Customizer}: Model fine-tuning and adaptation
            \item \textbf{NeMo Evaluator}: Model evaluation and testing
            \item \textbf{NeMo Guardrails}: Safety and quality assurance
        \end{itemize}
    \item \textit{Challenges}: 
        Developing targeted remediation strategies presents several challenges. The available training data consists of restricted labeled information because 495 production cases includes only 32 incorrect routing examples and 10 incorrect rephrasing instances. The learning process becomes more difficult because enterprise terminology and acronyms need specialized knowledge to understand their context. The model size requirements force developers to find an optimal point between performance and response time for maintaining system performance. The quality of synthetic data remains a problem because artificial examples need to exactly replicate actual user input and error behavior to achieve success.
    \item \textit{Learnings}: 
        The LLM-as-a-Judge approach delivered excellent results by accurately detecting routing errors at a rate of 77\%. The few-shot synthetic data generation method demonstrated excellent results because it needed only four to five examples to create high-quality training data. The domain-specific fine-tuning of smaller models produced results that were comparable to those of larger 70B models. The NVIDIA NeMo microservices stack modular design allowed developers to quickly test and optimize individual components which sped up the entire development cycle.
\end{enumerate}

\subsubsection{Execute Component (\textbf{E} in MAPE)}
\begin{enumerate}[label=\textit{\Roman*.}, leftmargin=3em]
    \item \textit{Problem}: The deployment of enhanced models to production requires various sequential operations which help reduce system downtime. The deployment of 70B parameter models leads to negative impacts on user experience and operational efficiency because they tend to have higher latency and cost.
    \item \textit{Solution}: Using NVIDIA NeMo Customizer, we executed model fine-tuning and progressive deployment:\\\\
        \textbf{Router Optimization Results:}
            \begin{itemize}
                \item Baseline: Llama 3.1 70B - 96\% accuracy, 0.26s latency
                \item Fine-tuned: Llama 3.1 8B - 96\% accuracy, 0.08s latency
                \item Achievement: 10x model size reduction, 70\% latency reduction
            \end{itemize}
    
        \textbf{Rephrasal Enhancement Results:}
            \begin{itemize}
                \item Baseline: Llama 3.1 70B - 73.8\% accuracy, 1.9s latency
                \item Fine-tuned: Llama 3.1 8B - 77.5\% accuracy, 1.1s latency
                \item Achievement: 3.7\% accuracy improvement, 40\% latency reduction
            \end{itemize}
    \item \textit{Challenges}: 
        The system faces major production risks because any unwanted changes will affect more than 30,000 users by degrading system performance. The system requires effective rollback mechanisms to perform fast updates and reduce system downtime during problematic changes. The system requires ongoing performance tracking to monitor change effects on different query domains while maintaining uniform quality standards. The deployment process requires teams to work together effectively because data scientists need to coordinate with engineers and operations staff to handle dependencies and preserve system stability.
    \item \textit{Learnings}: 
        The deployment process should include Canary and staged deployments to introduce changes to limited user groups before complete system deployment helps protect against unexpected system problems. The implementation of defined rollback procedures enables teams to safely return to previous updates when performance deterioration occurs. The monitoring of essential performance indicators including accuracy, latency and user feedback after deployment helps detect system deterioration at its beginning stages. The release process benefits from clear handoffs between data scientist, engineer and product manager which enables effective team collaboration. Users will develop more trust in new model versions when organizations maintain open communication about system updates.
\end{enumerate}

\section{Experimental Evaluation}

\subsection{Experimental Setup}

We evaluated our Data Flywheel implementation on NVIDIA's NVInfo bot through systematic feedback analysis and targeted improvements:

\begin{itemize}
\item \textbf{User Base}: 800 active users per week
\item \textbf{Feedback Dataset}: 1,224 human feedback samples from production (729 thumbs-up, 495 thumbs-down)
\item \textbf{Baseline Models}: Llama 3.1 70B for routing and query rephrasal
\item \textbf{Fine-tuning Models}: Llama 3.1 8B, Llama 3.2 3B/1B
\item \textbf{Infrastructure}: NVIDIA NeMo Customizer microservices for customization
\end{itemize}

\subsection{Error Analysis from User Feedback}

Through analysis of 495 negative feedback samples, we identified two primary failure modes:

\begin{table}[h]
\centering
\caption{Error Classification from User Feedback}
\label{tab:error_analysis}
\begin{tabular}{lcc}
\toprule
\textbf{Error Type} & \textbf{Count} & \textbf{Percentage} \\
\midrule
Routing Errors & 26/495 & 5.25\% \\
Rephrasal Errors & $\sim$16/495 & 3.2\% (extrapolated) \\
Other Errors & 453/495 & 91.5\% \\
\bottomrule
\end{tabular}
\end{table}

Example failures identified:
\begin{itemize}
\item \textbf{Routing Error}: "How many vacation days does NVIDIA Canada have?" was incorrectly routed to Holiday Expert instead of Policies Expert
\item \textbf{Rephrasal Error}: "What is the role of the RESS planning team?" failed due to incorrect expansion (RESS = Real Estate \& Site Services)
\end{itemize}

\subsection{Fine-Tuning Results}

\subsubsection{Infrastructure and Fine tuning Method}
To address key failure modes in the NVInfo RAG pipeline, we adopted LoRA via PEFT to optimize critical components such as routing and query rephrasal. LoRA enables targeted updates to transformer weights using lightweight, low-rank matrices, making it well suited for rapid iteration on curated failure samples without requiring full model retraining.

All fine-tuning was performed on an NVIDIA DGX Station equipped with 4× A100 GPUs (80 GB each), which provided the compute capacity needed for efficient parallel training on long-sequence, high-volume datasets.

\subsubsection{Expert Routing Optimization}

After gathering user feedback and incorporating subject-matter expert (SME) corrections, we compiled a curated dataset. In total, we collected 761 data points, consisting of 729 original samples and 32 additional corrections generated by the LLM-as-Judge. After removing duplicates, the dataset was reduced to 685 unique samples. For experimentation, we adopted a 60/40 split between training and testing sets.

\begin{table}[h]
\centering
\caption{Router Fine-Tuning Results: 10x Model Size Reduction}
\label{tab:router_results}
\begin{tabular}{lcc}
\toprule
\textbf{Model} & \textbf{Accuracy} & \textbf{Latency (s)} \\
\midrule
Llama 3.1 70B (baseline) & 96\% & 0.26 \\
Llama 3.1 8B (no tuning) & 14\% & 0.08 \\
Llama 3.1 8B + prompt-tuning & 86\% & 0.08 \\
Llama 3.1 8B + fine-tuning & \textbf{96\%} & \textbf{0.08} \\
Llama 3.2 3B + fine-tuning & 94\% & -- \\
Llama 3.2 1B + fine-tuning & 94\% & -- \\
\bottomrule
\end{tabular}
\end{table}

Key achievement: Maintained 96\% accuracy while reducing model size by 10x and latency by 70\%.

\subsubsection{Query Rephrasal Enhancement}

We manually analyzed 250 samples, from which we identified 10 candidates for rephrasing. To further expand the dataset, we generated 5,000 synthetic samples using the Llama 3.1 405B model with few-shot examples. For downstream experiments, the data was partitioned into an 80/10/10 split across training, validation, and test sets.

\begin{table}[h]
\centering
\caption{Query Rephrasal Fine-Tuning Results}
\label{tab:rephrasal_results}
\begin{tabular}{lcc}
\toprule
\textbf{Model} & \textbf{Accuracy} & \textbf{Latency (s)} \\
\midrule
Llama 3.1 70B (baseline) & 73.8\% & 1.9 \\
Llama 3.1 8B Fine-Tuned & \textbf{77.5\%} & \textbf{1.1} \\
\bottomrule
\end{tabular}
\end{table}

Key achievement: 3.7\% accuracy improvement with 40\% latency reduction and 10x model size reduction.

\subsection{Improvements Achieved Through the Data Flywheel}

Table \ref{tab:corrected_issues} shows examples of issues resolved through the data flywheel.

\begin{table*}[t]
\centering
\caption{Examples of Corrected Issues through Data Flywheel}
\label{tab:corrected_issues}
\begin{tabular}{p{3.5cm}p{3cm}p{3.5cm}p{4cm}}
\toprule
\textbf{User Query} & \textbf{Original Failure} & \textbf{After Fine-tuning} & \textbf{Result} \\
\midrule
"What is the role of the RESS planning team at NVIDIA?" & Rephrasal Error: Incorrectly expanded to "Resource Planning team" & Correct Rephrase: "NVIDIA RESS planning team role", "RESS planning team responsibilities" & The role of RESS (Real Estate and Site Services) Planning team is to manage site operations, support lease delivery... \\
\midrule
"How many vacation days does NVIDIA Canada have?" & Router Error: Sent to Holiday Expert instead of Policies Expert & Correctly routed to Policies Expert & According to the Canada Vacation Policy, employees receive... \\
\bottomrule
\end{tabular}
\end{table*}

\section{Discussion}

\subsection{Key Achievements}

\subsubsection{Model Size and Efficiency}
The model achieved a 10× reduction in size from 70B to 8B parameters while maintaining 96\% routing accuracy. The results show that domain-specific models with smaller sizes can achieve comparable performance to their larger general-purpose counterparts when fine-tuned properly. The model size reduction enabled a 70\% decrease in latency which made real-time applications more practical.

\subsubsection{Accuracy Improvements}
The 3.7\% improvement in query rephrasal accuracy shows significant advancement for production systems although it seems small at first glance. The combination of faster performance through reduced latency by 40\% and improved accuracy leads to enhanced user experience through quicker and more accurate system interactions.

\subsubsection{Systematic Error Identification}
The analysis of 495 feedback samples revealed that routing errors combined with rephrasal errors made up less than 10\% of all system failures at 5.25\% and 3.2\% respectively. The results indicate that the retrieval and ranking and generation stages of the RAG pipeline need further optimization efforts.

\subsection{Practical Challenges Encountered}

\subsubsection{Low Feedback Participation}
The system received feedback from 495 employees out of 30,000 users which shows difficulties in obtaining large-scale feedback data. The insufficient number of participants in the study creates sampling bias which reduces the generalizability of the obtained results. The system uses query reformulation as an additional data source but it does not replace the need for direct user feedback.

\subsubsection{Manual Analysis Bottleneck}
The process of reviewing 250 samples for rephrasal errors took too much time and proved to be unworkable at scale. The feedback pipeline faces a major bottleneck because LLM-as-Judge helps identify routing errors but there is no automated solution for this process.

\subsubsection{Privacy and Compliance}
The enterprise policies restricted storing complete query-response pairs which restricted thorough analysis of the data. The process of handling feedback data became more complicated because of PII removal requirements and GDPR and CCPA compliance regulations.

\subsubsection{Synthetic Data Generation}
The creation of 5,000 synthetic examples for rephrasal training proved successful. The process of maintaining high-quality and contextually accurate data required advanced prompt engineering techniques and validation procedures which raised operational costs for data augmentation.

\subsection{Implications for Enterprise AI}

\subsubsection{Proprietary Data as Differentiator}
The research proves that properly managed enterprise data through a data flywheel system creates a sustainable business advantage for organizations. Real-world usage data serves as the foundation for developing AI agents which become both adaptive and resilient.

\subsubsection{Importance of Modular Architecture}
NVIDIA NeMo tools consisting of Curator and Customizer and Evaluator and Guardrails proved essential for establishing the data flywheel system. The platform's modular design allowed separate optimization of individual components and fast development cycles which are essential for enterprise flexibility.

\subsubsection{TCO Reduction Through Model Optimization}
The model size reduction by 10 times delivered better performance while simultaneously reducing infrastructure expenses. Organizations that operate multiple AI agents can achieve substantial TCO reductions through optimization efforts which preserve service quality standards.

\subsection{Future Work}

\subsubsection{Automated Error Attribution}
The development of machine learning classifiers which can identify all RAG pipeline errors stands as the main objective. The system will achieve better scalability in root cause analysis through automated error classification which decreases human involvement in review processes.

\subsubsection{Continuous Learning Without Forgetting}
The ability to update models continuously while preserving current performance levels and preventing catastrophic forgetting remains vital for supporting incremental learning progress in changing enterprise environments.

\subsubsection{Multi-Agent Coordination}
The data flywheel concept should advance to support coordinated development between different specialized agents which represents a logical progression. System-wide intelligence and stable performance in complex enterprise systems depend on coordinated efforts between different system components.

\section{Conclusion}
The research presented real-world data flywheel implementation results for NVInfo Knowledge Assistant at NVIDIA which showed how enterprise AI agents can learn from their errors to improve their performance. The MAPE framework enabled us to convert user feedback into quantifiable performance enhancements which improved both model accuracy and user interaction quality.

Our main accomplishments included reducing model size by 10 times (from 70B to 8B parameters) while achieving 96\% routing accuracy and improving query rephrasal accuracy by 3.7\% and decreasing latency by 40\%. The experimental results confirm that optimized smaller models perform at least as well as larger models while reducing operational expenses.

The evaluation of 495 feedback samples showed that routing errors and rephrasal errors combined to make up only 8.45\% of total failure cases thus becoming the most suitable targets for optimization efforts. The targeted refinement approach using restricted training data produced substantial performance enhancements which proved the effectiveness of focused model improvement without needing extensive retraining.

The deployment process revealed two essential lessons about the challenges of obtaining large-scale feedback data and the restrictions that enterprise privacy regulations create. The solution we developed using implicit signal detection and synthetic data creation and privacy-friendly processing methods provides organizations with a functional method to handle their data collection and privacy restrictions.

The success of enterprise workflows in the future will depend on continuous AI agent improvement capabilities. Organizations that establish data flywheel systems will experience faster AI system development and operational efficiency and enhanced user satisfaction. Enterprise AI success depends on creating adaptive systems which learn from all user interactions rather than seeking flawless models at deployment. AI agents that receive feedback-based learning capabilities will develop into self-enhancing assets which improve their capabilities and value through continuous evolution.

\bibliographystyle{IEEEtran}
\bibliography{references}

\onecolumn
\appendix
\section{NVInfo AI Architecture}
\label{sec:nvinfo_architecture}
\begin{center}
    \textbf{NVInfo AI Architecture}
\end{center}

\begin{figure*}[ht!]
\centering
\includegraphics[width=0.85\textwidth]{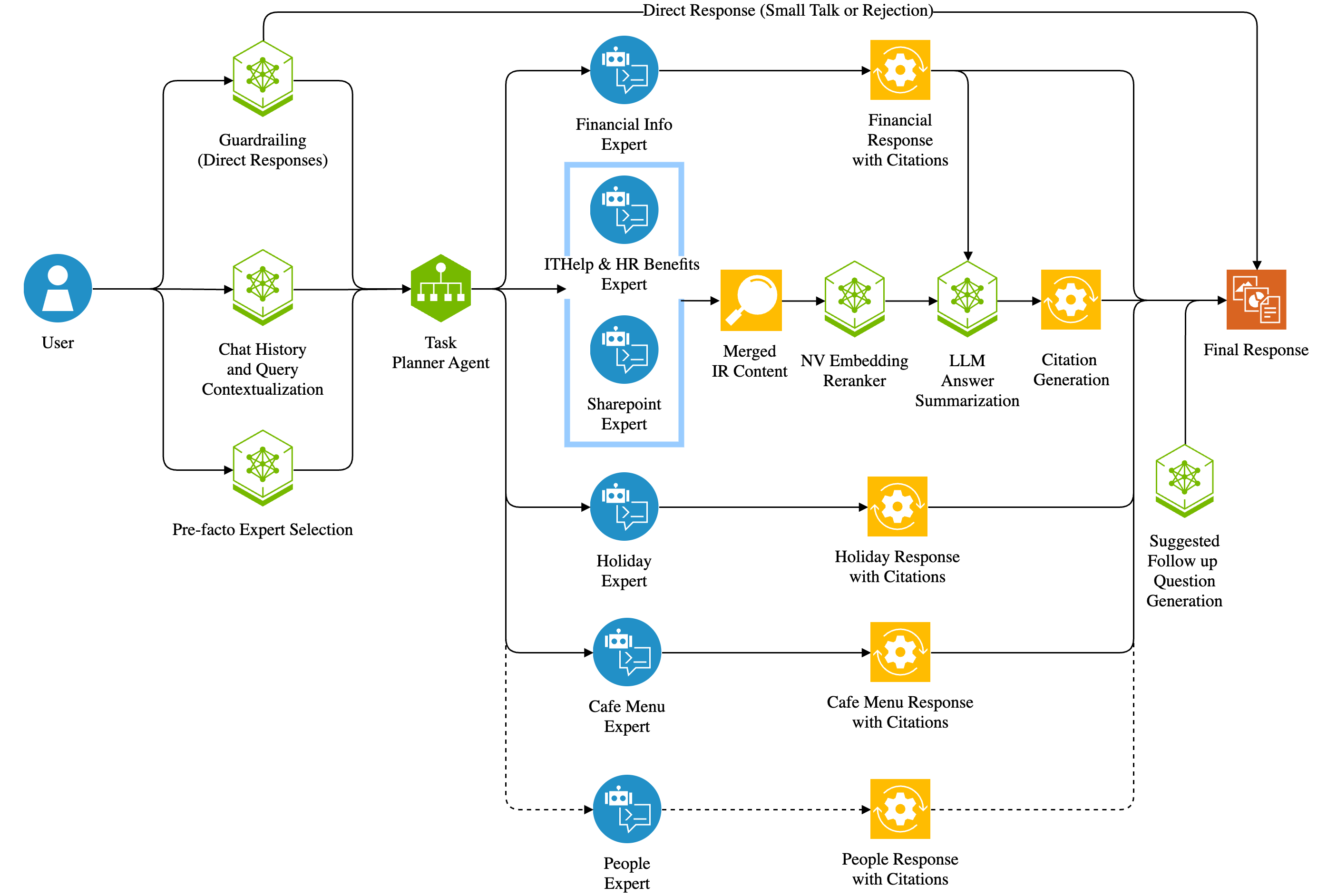}
\caption{NVInfo AI Mixture of Experts Architecture showing the complete RAG pipeline with Router, seven specialized domain experts, query rephrasing, retrieval, reranking, answer generation, and citation generation components}
\label{fig:nvinfo_architecture}
\end{figure*}

The architecture shown in Figure \ref{fig:nvinfo_architecture} illustrates the complete NVInfo AI system, which processes employee queries through a sophisticated pipeline:
\begin{itemize}
\item \textbf{Router}: Classifies incoming queries and routes them to the appropriate domain expert
\item \textbf{Seven Domain Experts}: Specialized models for Financial Info, IT Help \& HR Benefits, SharePoint, Holidays, Cafe Menu, People, and NVIDIA Policies
\item \textbf{Query Processing Pipeline}: Includes rephrasing, retrieval, reranking, answer generation, and citation generation
\item \textbf{Feedback Loop}: Captures user satisfaction signals for continuous improvement
\end{itemize}

\clearpage
\section{NVInfo Response and Feedback Capture Architecture}
\label{sec:nvinfo_data_capture}
\begin{center}
    \textbf{NVInfo Response and Feedback Capture Architecture}
\end{center}
\begin{figure*}[ht!]
\centering
\includegraphics[width=0.85\textwidth]{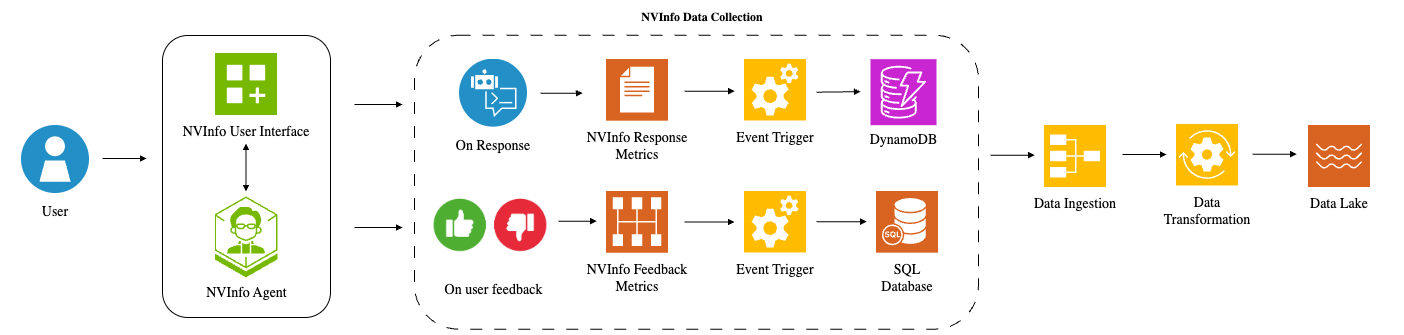}
\caption{NVInfo AI Response and Feedback Capture Architecture showing the complete data collection, ingestion and transformation components}
\label{fig:nvinfo_data_capture}
\end{figure*}
The figure \ref{fig:nvinfo_data_capture} illustrates the end-to-end data flow from user interaction with the NVInfo AI system to structured data storage for future system improvement. It highlights two main types of data captured(response metrics and user feedback metrics) and their subsequent processing.
\begin{itemize}
\item \textbf{User Interaction and Metrics Collection}: The data flow begins when a user interacts with the NVInfo User Interface, which connects to the NVInfo Agent, a domain-aware generative AI assistant that delivers structured, context-rich responses with citations. Each response is logged as part of NVInfo Response Metrics, capturing details such as query intent, routing, latency, and completeness. If the user provides feedback (e.g., thumbs up or down), NVInfo Feedback Metrics are recorded, including sentiment, error types, and optional comments. These metrics trigger events that stream response data to DynamoDB and feedback data to a SQL database, enabling structured downstream processing.
\item \textbf{Data Ingestion and Transformation}: A centralized data ingestion pipeline runs every 4 hours via a scheduled cron job to extract the latest response and feedback records from DynamoDB and SQL databases. This ensures timely synchronization while minimizing system load during peak usage periods.
\item \textbf{PySpark-based Data Transformation}: The ingested data is processed through a PySpark-based pipeline that performs cleaning, normalization, and enrichment. It maps feedback to specific conversation sessions, standardizes sentiment scores, and parses routing and rephrasal trace logs to identify failure modes. The resulting structured views capture model-side performance metrics such as routing accuracy and response latency, as well as user-side indicators like feedback sentiment and interaction quality, together providing a holistic picture of system effectiveness.
\item \textbf{Data Lake Storage}: The structured outputs are stored in a scalable data lake for long-term access and analysis. These views support downstream tasks such as dashboarding, fine-tuning, error analysis, and offline evaluation, contributing to continuous improvement of the NVInfo Agent.
\end{itemize}

\clearpage
\section{RAG System Failure Points}
\label{sec:rag_failures}
\begin{center}
    \textbf{RAG System Failure Points}
\end{center}
\begin{figure*}[ht!]
\centering
\includegraphics[width=0.85\textwidth]{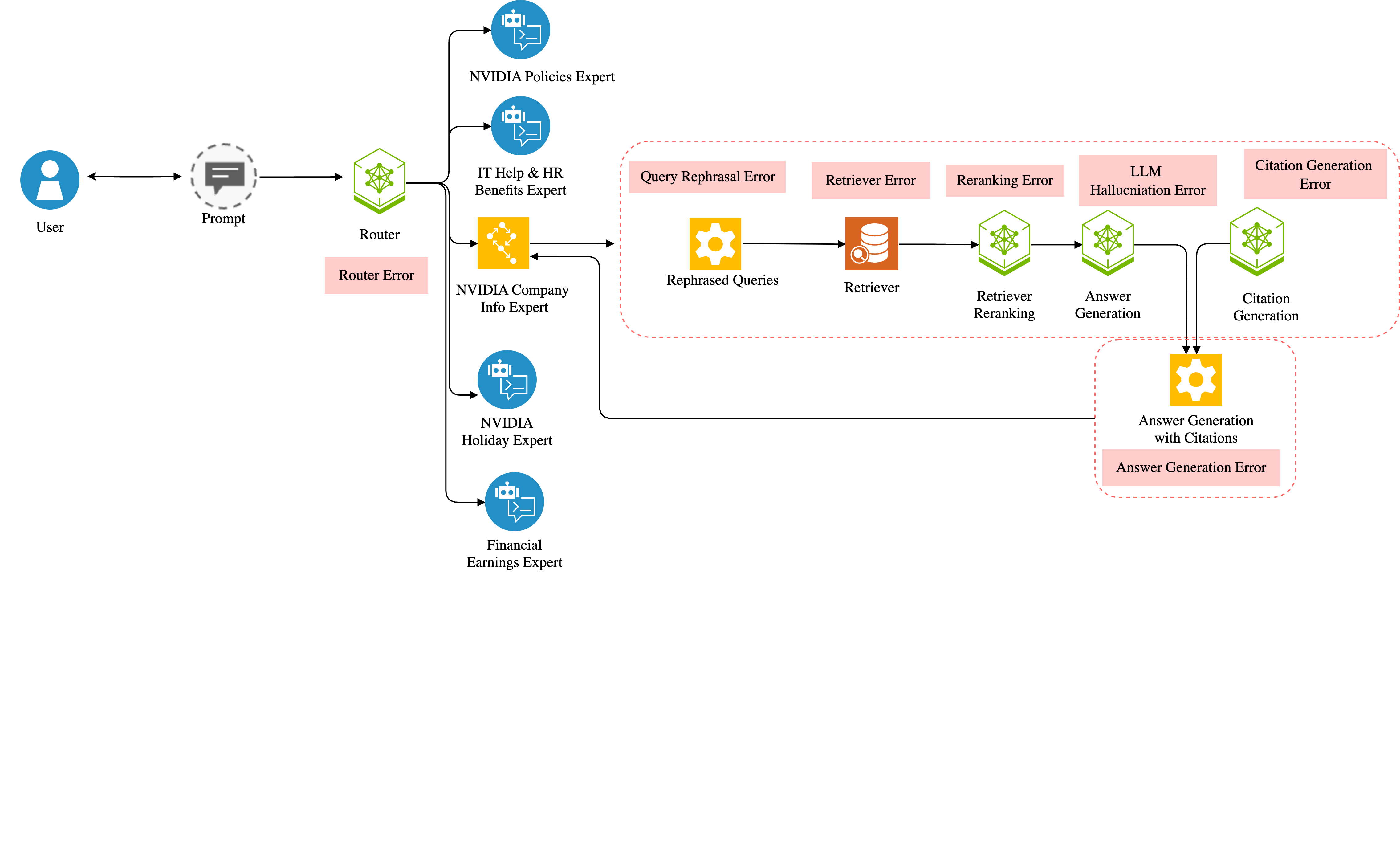}
\caption{Sequential failure points in the RAG pipeline from query routing to answer generation}
\label{fig:rag_failures}
\end{figure*}

The RAG pipeline, as shown in Figure \ref{fig:rag_failures}, faces challenges at each processing stage. These failure points were identified through analysis of 495 negative feedback samples collected over 3 months:
\begin{itemize}
\item \textbf{Router - Query Understanding}: Misclassification of user intent (5.25\% of failures)
\item \textbf{Query Rephrasing Error}: Incorrect query expansion (3.2\% of failures)
\item \textbf{Retriever Error}: Failure to find relevant documents despite their existence
\item \textbf{Reranking Error}: Incorrect prioritization of retrieved documents
\item \textbf{LLM Hallucination}: Generation of plausible but incorrect information
\item \textbf{Citation Generation Error}: Incorrect or missing source attribution
\item \textbf{Answer Generation Error}: Poor synthesis of retrieved context
\end{itemize}

\clearpage
\section{NVInfo Interface Examples}
\label{sec:interface_examples}
\begin{center}
    \textbf{NVInfo Interface Examples}
\end{center}
\begin{figure*}[h]
\centering
\begin{subfigure}[b]{0.48\textwidth}
    \includegraphics[width=\textwidth]{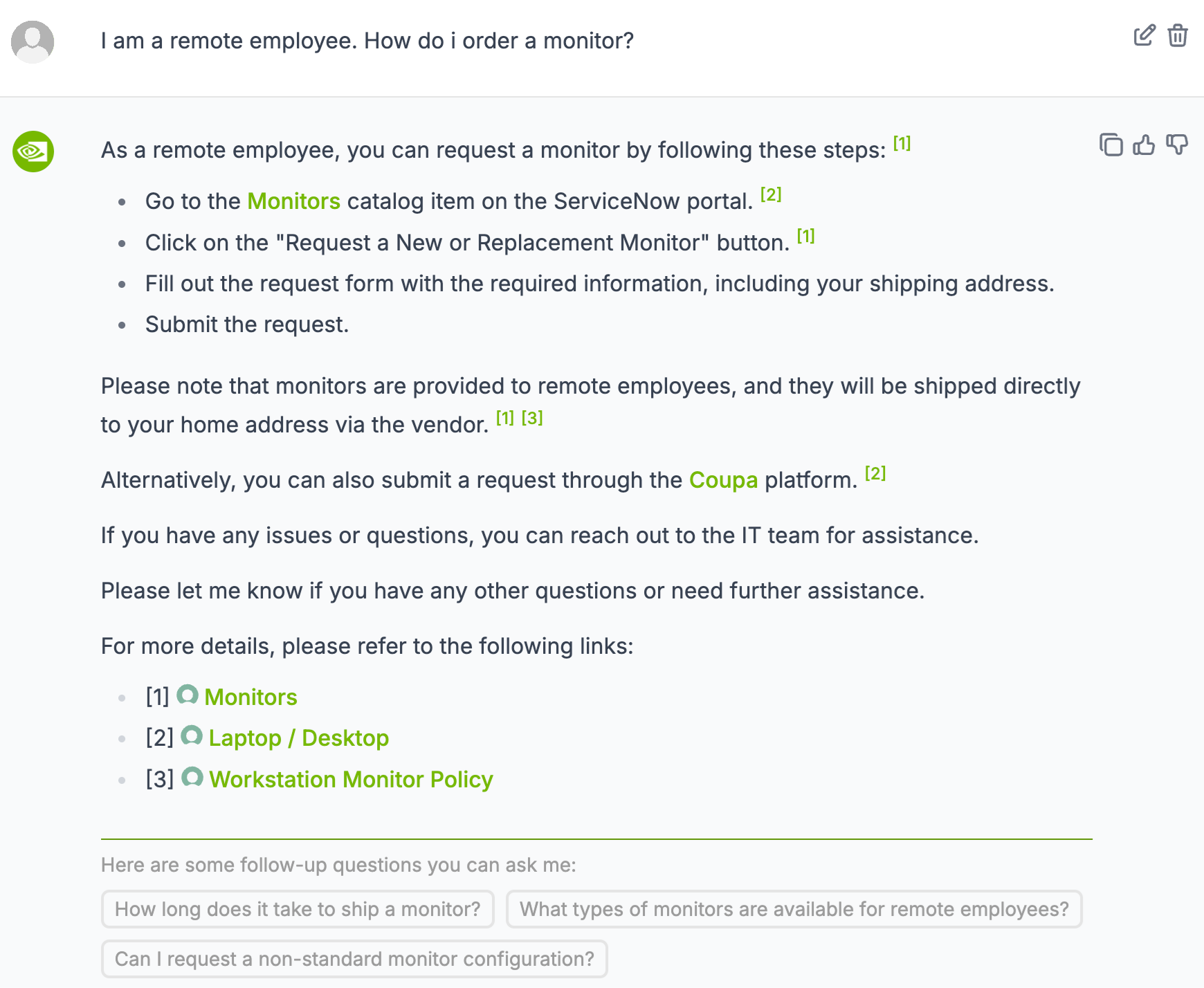}
    \caption{IT Help Expert providing technical support for WFH monitor ordering}
    \label{fig:it_help_appendix}
\end{subfigure}
\hfill
\begin{subfigure}[b]{0.48\textwidth}
    \includegraphics[width=\textwidth]{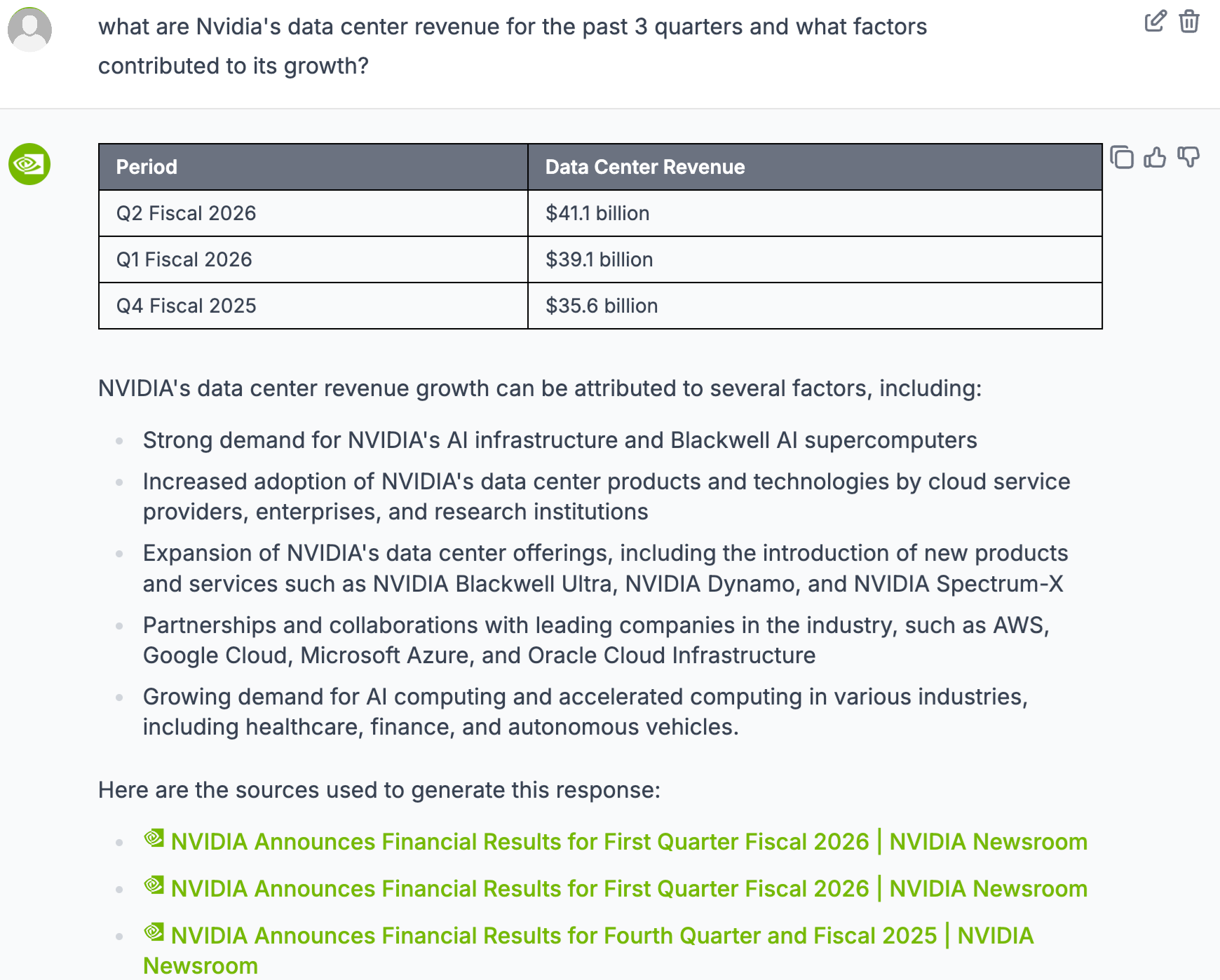}
    \caption{Financial Earnings Expert analyzing quarterly revenue growth}
    \label{fig:financial_appendix}
\end{subfigure}
\begin{subfigure}[b]{0.48\textwidth}
    \includegraphics[width=\textwidth]{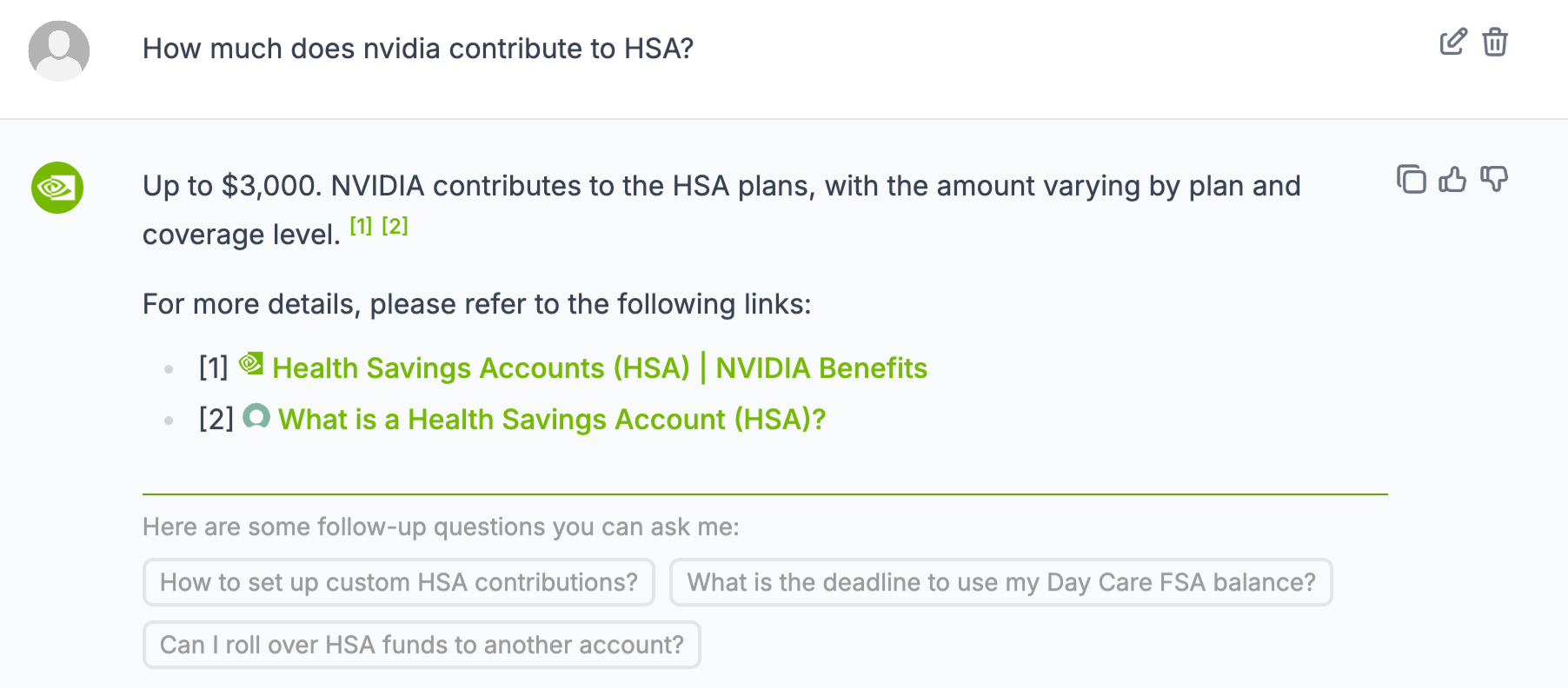}
    \caption{HR Benefits Expert explaining HSA contribution details}
    \label{fig:hr_benefits_appendix}
\end{subfigure}
\hfill
\begin{subfigure}[b]{0.48\textwidth}
    \includegraphics[width=\textwidth]{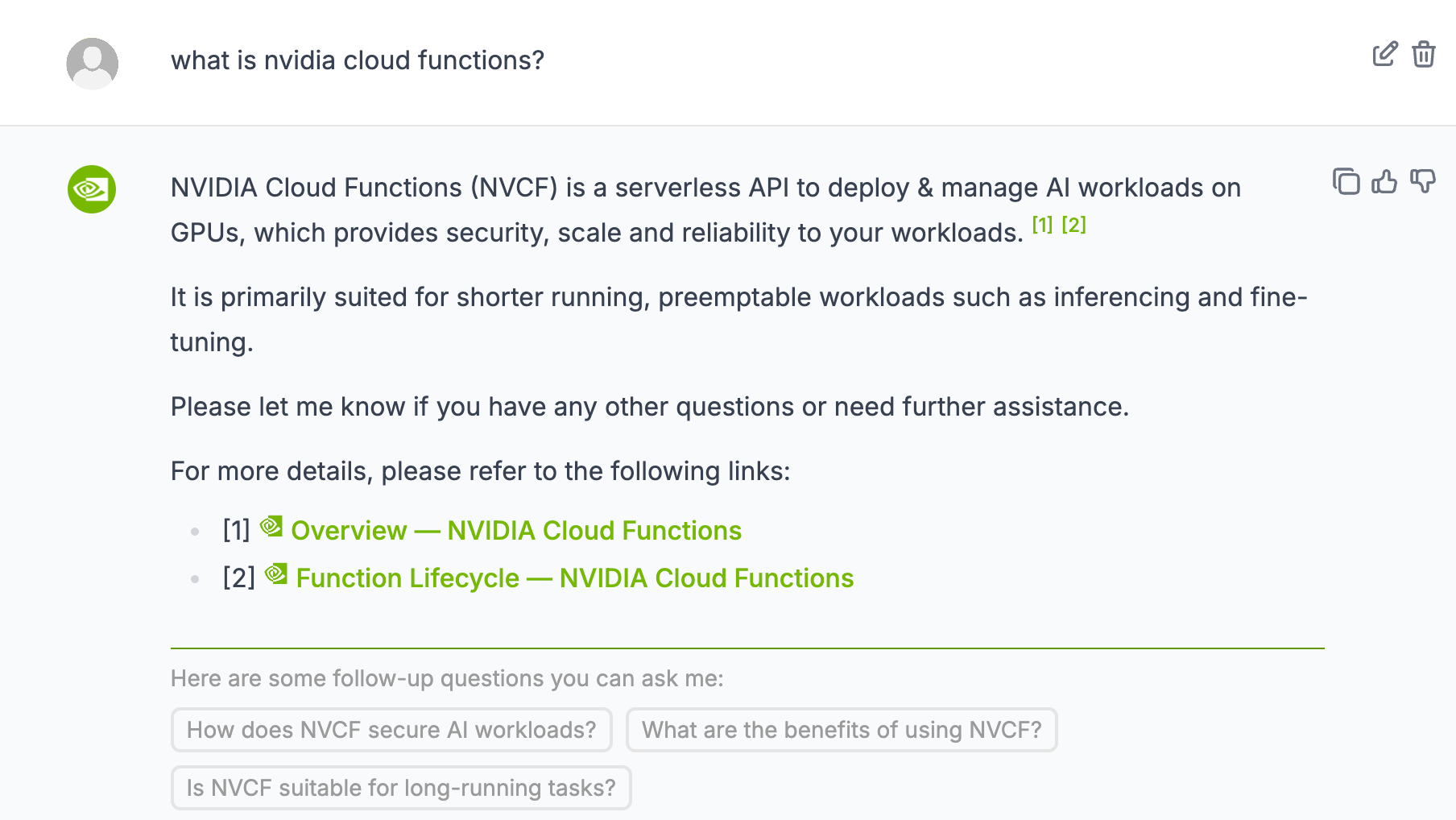}
    \caption{Company Info Expert providing NVC function information}
    \label{fig:product_doc_appendix}
\end{subfigure}
\caption{Representative NVInfo AI interface examples showing mixture-of-experts responses across different enterprise domains}
\label{fig:nvinfo_examples_appendix}
\end{figure*}

The interface examples above demonstrate the system's capabilities:
\begin{itemize}
\item \textbf{IT Support (Fig. \ref{fig:it_help_appendix})}: Shows step-by-step guidance for ordering WFH equipment through the ServiceNow portal
\item \textbf{Financial Analysis (Fig. \ref{fig:financial_appendix})}: Provides quarterly revenue data with year-over-year growth metrics and detailed breakdowns
\item \textbf{HR Benefits (Fig. \ref{fig:hr_benefits_appendix})}: Displays HSA contribution tables with employer matching details for different fiscal quarters
\item \textbf{Product Documentation (Fig. \ref{fig:product_doc_appendix})}: Explains technical concepts like NVIDIA Cloud Functions with architecture overview
\end{itemize}

\clearpage
\section{Prompt for router error LLM-as-a-judge classification}
\label{sec:router_error_prompt}
\begin{center}
    \textbf{Prompt for router error LLM-as-a-judge classification}
\end{center}
\FloatBarrier 
\noindent
\begin{lstlisting}[style=prompt,caption={Prompt for router error LLM-as-a-judge classification (complete example)},label={fig:router_error_prompt_all}]
Question: How do I submit a referral?
Tools: ['it_benefits_help', 'nvinfo_policies_expert']
Reasoning: This question is related to NVIDIA policy which means it should be sent to either 'it_benefits_help' or 'nvinfo_policies_expert'.
Answer: YES

Question: When can I sign up for a new health plan?
Tools: ['finance_expert']
Reasoning: This question is related to employee benefits which means it should be sent to 'it_benefits_help' instead of 'finance_expert'.
Answer: NO

Question: what was NVIDIA's Q3 revenue in fiscal 2024?
Tools: ['finance_expert']
Reasoning: This question is related to NVIDIA's earnings which means it should go to 'finance_expert'.
Answer: YES

Question: Is Mercedes Benz using NVIDIA's digital twin technology?
Tools: ['it_benefits_help', 'nvinfo_policies_expert']
Reasoning: This question is related to NVIDIA's products and therefore should have gone to 'it_benefits_help'.
Answer: YES

Question: What is the vacation policy at NVIDIA?
Tools: ['nvinfo_holiday_expert']
Reasoning: This question is related to NVIDIA policy which means it should be sent to either 'it_benefits_help' or 'nvinfo_policies_expert'.
Answer: NO

Question: When is the next free day at NVIDIA?
Tools: ['nvinfo_holiday_expert']
Reasoning: The user is trying to find the date of a holiday which means that the question should be sent to ['nvinfo_holiday_expert'].
Answer: YES

Question: When is the first open stock sale period in 2025?
Tools: ['finance_expert']
Reasoning: This question is related to NVIDIA's company finances and should therefore be sent to 'finance_expert'.
Answer: YES

Question: How many unused vacation days can I carry over?
Tools: ['it_benefits_help', 'nvinfo_policies_expert']
Reasoning: This question is related to NVIDIA policy and employee benefits which means it should be sent to either 'it_benefits_help' or 'nvinfo_policies_expert'.
Answer: YES

Question: Who heads up wwfo?
Tools: ['finance_expert']
Reasoning: This question is related to NVIDIA's leadership which means that it should be sent to 'finance_expert'.
Answer: YES

Question: Who is John Smith?
Tools: ['finance_expert']
Reasoning: The user is trying to find information about a specific person which means that this question should go to 'it_benefits_help' or 'nvinfo_policies_expert'.
Answer: NO

Question: What are the latest hardware offerings by nvidia?
Tools: ['it_benefits_help', 'nvinfo_policies_expert']
Reasoning: This question is related to NVIDIA's products and therefore should have gone to 'it_benefits_help'.
Answer: YES

Question: What is gb200 nvl72?
Tools: ['finance_expert']
Reasoning: This question is related to NVIDIA's products and therefore should have gone to 'it_benefits_help'.
Answer: NO

Question: When will the 2025 free days be officially announced?
Tools: ['it_benefits_help', 'nvinfo_policies_expert']
Reasoning: This question is related to NVIDIA's policies or benefits, so it should be sent to 'it_benefits_help' or 'nvinfo_policies_expert'.
Answer: YES

Question: Does nvidia offer financial advice services?
Tools: ['finance_expert']
Reasoning: This question is related to NVIDIA's policies or benefits, so it should be sent to 'it_benefits_help' or 'nvinfo_policies_expert'.
Answer: NO

Question: What was the year-over-year (YoY) and quarter-over-quarter (QoQ) growth for Q2 Fiscal 2025?
Tools: ['finance_expert']
Reasoning: This question is related to NVIDIA's earnings and should therefore be routed to 'finance_expert'.
Answer: YES

Question: How do I order a mouse?
Tools: ['it_benefits_help', 'nvinfo_policies_expert']
Reasoning: This question is related to procuring a work accessory, which means that it should go to either 'it_benefits_help' or 'nvinfo_policies_expert'.
Answer: YES

Question: I'm getting a VPN error
Tools: ['finance_expert']
Reasoning: This question is related to an IT issue, which means that it should go to either 'it_benefits_help' or 'nvinfo_policies_expert'.
Answer: NO

QUERY: {query}
TOOLS: {experts}
\end{lstlisting}
\FloatBarrier

\clearpage
\section{Regression dataset}
\label{sec:regression_data}
\begin{center}
    \textbf{Regression Dataset}
\end{center}

The NVInfo regression dataset is actively curated and regularly updated, currently comprising around 300 queries that cover a range of domains including NVIDIA benefits, holidays, company policies, and IT Help. Each query in the dataset contains the corresponding ground truth and expected citation values. LLM-as-judge framework is leveraged to evaluate the quality of NvInfo generated answers against the regression dataset. The criteria for judgment is based on metrics such as correctness, helpfulness, and conscientiousness.

\clearpage
\section{Synthetic data generation}
\label{sec:synthetic_data_generation}
\begin{center}
    \textbf{Synthetic data generation}
\end{center}

As part of our continuous improvement efforts, we identified that refining the way queries are phrased could significantly enhance retrieval accuracy. Given that our retrieval system is highly sensitive to keyword usage, an in-depth analysis of existing feedback was conducted. During this analysis, we discovered that decomposing certain queries into sub-queries improved the recall of relevant information from our internal retrieval systems. This, in turn, led to more accurate and contextually appropriate responses.
To tackle this, we conducted a detailed review of about 250 examples from our "thumbs down" feedback dataset, focusing specifically on queries related to our SharePoint expert system. We noticed that some queries weren’t retrieving the most relevant information due to a lack of understanding of NVIDIA-specific acronyms or context.
By manually rephrasing these queries, we found that we could significantly improve the retrieval of the intended information. From an initial set of 250 examples, we identified 10 key candidates showcasing common patterns of misinterpretation or context loss. For instance, the query "I am based in the netherlands, when is pay day?" was initially rephrased as "payday schedule united states, employer pay dates usa." We manually corrected this to "payday schedule netherlands" and "netherlands pay days."
To extend this improvement beyond the feedback dataset, we integrated these rephrased examples into our synthetic data generation pipeline. We leveraged these examples as few-shot prompts for our large language model. By providing SharePoint-related website content as context, we instructed the LLM to generate both original and rephrased queries for all documents. This method allowed us to produce approximately 5,000 rephrased queries, thereby enriching our dataset and facilitating more effective fine-tuning of the agent. This focused enhancement significantly improved the SharePoint expert's ability to retrieve and deliver the most relevant information with increased accuracy.

\begin{promptcard}[title={Prompt for Synthetic Data Generation}]
You are a data annotator generating \textbf{questions}, \textbf{answers}, and \textbf{rephrased questions} from an input document and its URL.

\promptsubtitle{Guidelines}
\begin{itemize}
  \item Identify key phrases and entities in the document and generate questions around them.
  \item Generate questions answerable using information contained in the \emph{input document}.
  \item Do \emph{not} write questions that require viewing the document to understand the question.
  \item Avoid phrases like “according to the document/author”, “in this document”, etc.
  \item Questions may also be key phrases found in the document.
  \item Ensure the document contains the complete answer to your question.
  \item Provide enough context in the question to lead to the specific answer in the document.
  \item Vary phrasing, vocabulary, complexity, and type of questions.
  \item \textbf{Do not} copy exact phrasing; use your own words.
  \item Prefix questions with \texttt{Question:} and answers with \texttt{Answer:}.
  \item Rephrase each question at least twice (query decomposition/expansion) to aid search.
  \item Final output \textbf{must} be a Python list.
  \item Rephrased queries are short, concise keyword/entity mixes; you may replace \texttt{nvidia} with \texttt{employer} or \texttt{company}.
  \item Provide two or more rephrased queries preserving intent and timeframe.
  \item If the question asks for “the next \textit{X} date” without time context, append \texttt{YYYY} (current or next year) in rephrased queries.\\
        \textit{Example:} Question: “when is the next NTech conference” $\rightarrow$
        “upcoming ntech 2024”, “ntech dates 2024”, “ntech schedule 2025”.
\end{itemize}

\promptsubtitle{Use the EnterpriseKnowledge tool when}
The user asks for non-sensitive information such as organization info, direct reports, phone numbers, benefits alternate ID, email addresses, working addresses, tax explanations, updating SSN instructions, or stock trading policies.

\promptsubtitle{Your action format MUST be}
\begin{lstlisting}[style=jsonstyle]
Thought: Provide a short analysis of your understanding from the Question.
Process: I need to use the Enterprise Knowledge tool
Action: EnterpriseKnowledge
Action Input: A single line Python list of rephrased queries MUST be generated.
\end{lstlisting}

\promptsubtitle{Strict JSON schema (return nothing else)}
\begin{lstlisting}[style=jsonstyle]
{
  "type": "object",
  "properties": {
    "Question": {
      "type": "string",
      "description": "Generated Question from the input document."
    },
    "Answer": {
      "type": "string",
      "description": "Corresponding Answer from the input document that answers the Question."
    },
    "Thought": {
      "type": "string",
      "description": "Short analysis of your understanding from the Question."
    },
    "Process": {
      "type": "string",
      "description": "I need to use the Enterprise Knowledge tool."
    },
    "Action": {
      "type": "string",
      "description": "EnterpriseKnowledge"
    },
    "Action Input": {
      "type": "list",
      "description": "A single line Python list of rephrased queries."
    }
  }
}
\end{lstlisting}

\promptsubtitle{Examples}
\textbf{Input Document:} \texttt{<Content of input document>} \\
\textbf{Input Document url:} \texttt{<url of input document>} \\[0.25em]
\textbf{Output}
\begin{lstlisting}[style=jsonstyle]
{
  "Question": "I am based in the Netherlands, when is pay day?",
  "Answer": "25th of every month",
  "Thought": "Payroll timing question; include location keywords in rephrased queries.",
  "Process": "I need to use the Enterprise Knowledge tool",
  "Action": "EnterpriseKnowledge",
  "Action Input": [
    "payday schedule netherlands",
    "netherlands pay days"
  ]
}
\end{lstlisting}

\textbf{Input Document:} \texttt{<Content of input document>} \\
\textbf{Input Document url:} \texttt{<url of input document>} \\[0.25em]
\textbf{Output}
\begin{lstlisting}[style=jsonstyle]
{
  "Question": "point me to gpu fcv page?",
  "Answer": "https://nvidia.sharepoint.com/sites/TechnicalTraining/ASIC%20teams.aspx",
  "Thought": "Needs GPU FCV (Full Chip Verification) page.",
  "Process": "I need to use the Enterprise Knowledge tool",
  "Action": "EnterpriseKnowledge",
  "Action Input": [
    "gpu fcv page company",
    "fcv gpu url"
  ]
}
\end{lstlisting}

\textbf{Input Document:} \texttt{<Content of input document>} \\
\textbf{Input Document url:} \texttt{<url of input document>} \\[0.25em]
\textbf{Output}
\begin{lstlisting}[style=jsonstyle]
{
  "Question": "ok, i'm looking for an nvidia icon for biotech / pharmaceuticals to use in a presentation. can you help me find that?",
  "Answer": "https://nvidia.sharepoint.com/sites/nvinfo/brand/Pages/default.aspx",
  "Thought": "Needs a company icon for biotech/pharma use.",
  "Process": "I need to use the Enterprise Knowledge tool",
  "Action": "EnterpriseKnowledge",
  "Action Input": [
    "company icons",
    "company logos biotech"
  ]
}
\end{lstlisting}

\promptsubtitle{Task output format}
Generate \textbf{3 pairs} by following the instructions based on the Input Document. \\
\textbf{Strictly return only a Python list of pairs and nothing else.}\\[0.25em]
\textbf{Input Document:} \texttt{<Content of input document>} \\
\textbf{Input Document url:} \texttt{<url of input document>} \\
\textbf{Output:} \texttt{\#\#\#}
\end{promptcard}

\end{document}